\documentclass[cameraready]{Interspeech}
\usepackage{multirow}
\usepackage{booktabs}
\usepackage{CJKutf8}

\title{Montreal Forced Aligner and the state of speech-to-text alignment in 2026}

\author[affiliation={1,2}, orcid=0000-0001-9941-1421, correspondingauthor]{Michael}{McAuliffe}
\author[affiliation={3}, orcid=0009-0003-3400-785X]{Kaylynn}{Gunter}
\author[affiliation={2}, orcid=0000-0003-2953-106X]{Michael}{Wagner}
\author[affiliation={2}, orcid=0000-0001-7675-2370]{Morgan}{Sonderegger}

\address{
    $^1$ University of Wisconsin--Madison, USA \\
    $^2$ McGill University and Centre for Brain, Language, and Music, Canada \\
    $^3$ University of Oregon, USA
}

\email{mcauliffe3@wisc.edu, kgunter@uoregon.edu, chael@mcgill.ca, morgan.sonderegger@mcgill.ca}


\usepackage{comment}

\begin{document}

\maketitle

\begin{abstract}

    The Montreal Forced Aligner (MFA) was released in 2016 and has since become the most widely used tool for forced alignment in research and industry.  In the decade since, MFA has undergone substantial development, including expanded coverage across more languages and dialects using larger open-source datasets, harmonized IPA dictionaries, model adaptation, cross-language phone remapping, and support utilities.  This paper documents MFA 3.0's developments since version 1.0 and evaluates MFA's performance across English, Japanese, and Korean, benchmarked against classic and neural forced aligners.    MFA 3.0 achieves state-of-the-art or near state-of-the-art performance across all four benchmark datasets  with mean boundary errors below 15 ms.  Adaptation and cross-language remapping are effective for languages outside MFA's training distribution, and pronunciation probability modeling and phonological rules provide gains in specific conditions.
    
\end{abstract}

\section{Introduction}

Forced alignment, the automatic temporal alignment of words and phonemes to a speech recording given its orthographic transcription, has become a standard first step in language science research across (socio)phonetics, language documentation, and psycholinguistics. Many forced aligners have been developed over the past 20 years (e.g., \cite{yuan2008speaker, rosenfelder2011fave, kisler2012signal, gorman2011prosodylab}), and the field has a healthy ecosystem of tools for different use cases. The Montreal Forced Aligner \cite{mcauliffe_montreal_2017}, released in 2016, has become the most widely used of these. In the intervening years, MFA has undergone substantial development, and the field has changed considerably: new aligners using neural ASR architectures have been proposed, larger and more diverse training datasets have become available, and the range of languages and use cases has expanded. Despite this, no systematic evaluation of current MFA against the range of available aligners has been published since its initial release; existing comparisons of forced aligners consider a small number of aligners and/or are restricted to English, and use different datasets and metrics (e.g., \cite{mahr2021performance,rousso_tradition_2024,kelley2024mason}).

In tandem with advances in the field, MFA has expanded its core functionality to leverage new data and tools and address common bottlenecks in language research pipelines, including incorporating libraries for transcription, tokenization, and corpus creation. The goal of this paper is two-fold: to document key developments in MFA since version 1.0, and to provide a systematic benchmark of MFA 3.0 against a range of currently available aligners across three languages. We evaluate MFA's performance on detecting word and phone boundaries in laboratory and conversational speech, and assess the contribution of MFA-specific functionality---adaptation, cross-language remapping, pronunciation probability modeling, and phonological rules---to alignment performance. MFA 3.0 achieves state-of-the-art performance across all three languages, outperforming classical and neural aligners in most comparisons, with mean boundary errors consistently below 15 ms.



\section{Background}

Development of MFA from 1.0 to 3.0 has been driven by rapid expansion in the \emph{use cases} of forced aligners in scientific research and the \emph{data and tools} available. We briefly cover each to motivate the features of MFA 3.0 described in Sec.~\ref{sec: mfa3}.

\subsection{Use cases}



A forced aligner consists minimally of an \emph{acoustic model} and \emph{pronunciation dictionary}; ten years ago aligners were applied to a few high-resource languages\footnote{Except the  Munich AUtomatic Segmentation system (MAUS), which by 2012 supported 7 languages \cite{kisler2012signal}.} with fixed broad-transcription dictionaries assuming a standard dialect (e.g.\ \cite{kiesling_variation_2006, yuan2008speaker,gorman2011prosodylab,rosenfelder2011fave}).

Forced aligners are now used across a much wider range of languages, dialects, and speaker populations, including low-resource and endangered languages \cite{dicanio2013using, johnson2018forced, babinski2019robin, ahn2024use, tosolini2025multilingual}, diverse dialects \cite{mackenzie2020assessing, fromont2023maximizing}, child speech \cite{knowles2018examining, mahr2021performance, christodoulidou2025semi}, and L2 speech \cite{williams2024analysis}.
%
%
%
%
%
This has gone hand-in-hand with expansion in the number of languages covered out-of-the-box by MFA (Table~\ref{tab:training_data}) as well as other aligners \cite{kisler2012signal,bigi2012sppas,rehman2025bfa}, meaning the software has a pretrained acoustic model and either a pronunciation dictionary or a grapheme-to-phoneme (G2P) tool.  

This expanded range of use cases has made 
how best to train and deploy forced aligners for target data outside their training data its own research area (e.g. \cite{ahn2024use,tosolini2025multilingual,christodoulidou2025semi,berez2023recent, coto2022computational,chodroff2025comparing}), and three broad strategies have emerged: 
(1) Using a pretrained aligner as-is, relying on acoustic model robustness. (2)
Adapting a pretrained aligner to the target language, dialect, or population via acoustic model adaptation or phone set remapping.
(3) Training a new aligner on target domain data.

This literature has found each strategy is effective in different cases, but in general training data quantity and diversity trumps language-specificity, with models trained on the largest datasets (e.g.\ MFA 3.0 Global-English)  competitive with or better than smaller language/dialect-specific ones. Strategies (2) and (3) are enabled by easily \emph{trainable} aligners, beginning with the ProsodyLab aligner \cite{gorman2011prosodylab}, the precursor to MFA.  The three strategies correspond to the three  features of MFA 3.0 evaluated in this paper's experiments: improved pretrained models, adaptation and remapping, and training from scratch.


Use cases have also expanded in terms of phonetic detail, reflected in the pronunciation dictionary. Demand has grown for narrower transcription reflecting dialectal forms and probabilistic pronunciation modeling, in parallel with forced aligners being increasingly used to study phonetic variation by choosing between variants of the same word \cite{schiel1998probabilistic,yuan2009investigating, schuppler2011acoustic, ryant2016large, wu2022extracting}. This has motivated a broader range of pronunciation dictionaries across forced aligners, with MFA in particular transitioning to an IPA phone set and trainable pronunciation probabilities (Sec.~\ref{sec:pron_dict}).

\subsection{Data and tools}

The data and tools available for building forced aligners have expanded substantially since MFA 1.0. Large open-source speech datasets---such as CommonVoice \cite{ardila_common_2020}, Multilingual LibriSpeech \cite{Pratap2020MLSAL}, and many others on OpenSLR and elsewhere---have made it feasible to train acoustic models on hundreds to thousands of hours across many languages in MFA 3.0, orders of magnitude more than was available 10 years ago. 
Crowd-sourced pronunciation resources such as WikiPron \cite{lee_massively_2020} and better grapheme-to-phoneme (G2P) systems \cite{gorman2016pynini,mortensen2018epitran,priva2021cross} have greatly expanded cross-linguistic pronunciation coverage. At the same time, the increase in data quantity has come with increased variability in quality, driving development of tooling for checking and correcting datasets: a key part of MFA development described in Sec.~\ref{sec:support}, and other speech database systems \cite{fromont2012labb, winkelmann2017emu}. Open-source tools for diarization (e.g.\ Pyannote \cite{bredin_pyannote_2020}) and transcription (e.g.\ WhisperX \cite{bain_whisperx_2023}, SpeechBrain \cite{ravanelli_speechbrain_2021}) have enabled increasingly automated end-to-end corpus processing and phonetic analysis pipelines (e.g.\ \cite{ahn2022voxcommunis,coats2023pipeline}), motivating MFA's integrations with these tools (Sec.~\ref{sec:support}).

\subsubsection{Neural architectures}


Forced aligners ten years ago were developed using classic HMM-GMM systems built with HTK \cite{young1993htk}, Julius \cite{lee2001julius}, or Kaldi \cite{Povey_ASRU2011}, where phone and word boundaries are a by-product of the ASR process. The field of ASR has since seen dramatic advances through neural architectures, with end-to-end models achieving much lower word error rates. Some neural ASR systems output word boundaries and can be used as forced aligners, but are not optimized for fine temporal boundary placement: with no pronunciation dictionary and no intermediate phone-level representation, they are trained to produce correct output strings rather than precise frame-level alignments, and CTC-based systems in particular explicitly collapse across different alignments for the same string \cite{rousso_tradition_2024}. We replicate and extend \cite{rousso_tradition_2024}'s comparison of MFA with neural ASR-based aligners using a wider  set of  aligners, including NeMo \cite{rastorgueva_nemo_2023}.

Recently, neural systems have been developed specifically for the forced alignment or phonetic segmentation task, including the Mason-Alberta Phonetic Segmentor (MAPS: \cite{kelley2024mason}), Charsiu \cite{zhu2022charsiu}, and the Bournemouth Forced Aligner (BFA: \cite{rehman2025bfa}), achieving competitive phone-level alignment performance. We include these in our evaluation (Sec.~\ref{sec:results}).

\section{Montreal Forced Aligner 3.0}
\label{sec: mfa3}

MFA is an open-source command line utility with prebuilt executables for Windows, Mac OSX, and Linux  \cite{mcauliffe_montreal_2017}. MFA 3.0 extends version 1.0 in four main areas. First, it leverages the increase in available data to provide an expanded set of pretrained acoustic models with greater coverage of languages and linguistic/social variation therein (Sec.~\ref{sec:acoustic model training}). Second, for use cases where out-of-the-box support is unavailable or insufficient, MFA 3.0 includes utilities for adapting acoustic models to novel languages and populations (Sec.~\ref{sec:adaptation}). Third, it integrates with open-source libraries to facilitate corpus creation and iteratively improving alignments (Sec.~\ref{sec:support}). Finally, MFA 3.0 emphasizes usability and iterative refinement as core design principles, with documentation and utilities to support each stage of the alignment pipeline (Sec.~\ref{sec:usability}).

\subsection{Pretrained models}
\label{sec:acoustic model training}


\subsubsection{Training data}

\label{sec:training_data}

Pretrained models and dictionaries for MFA 1.0 were largely based on GlobalPhone corpora \cite{schultz_globalphone_2013}, which range from 8 hours of speech to 30 hours of speech, with the exception of English which was trained on 1,000 hours of LibriSpeech \cite{panayotov_librispeech_2015}. Pretrained models for MFA 3.0 leverage large open-source speech datasets such as CommonVoice \cite{ardila_common_2020} and datasets on OpenSLR (e.g.\ Multilingual LibriSpeech \cite{Pratap2020MLSAL}) to train on data several orders of magnitude larger than 1.0, as seen in Table~\ref{tab:training_data}. 
There is also an increase in the coverage of variation. Datasets in MFA 1.0 primarily included data from single dialects (e.g., only Brazilian Portuguese), while MFA 3.0 includes multiple dialects (e.g., Brazilian and European Portuguese) for broader coverage in pronunciation variability. While several datasets have explicit information about dialect areas, others are unspecified and contain a mix of dialect variation within the dataset, including occasionally second language (L2) speakers. In addition to dialect coverage, these datasets also cover a range of speech styles (read and conversational), recording environments, and noise. Corpora used to train each MFA 3.0 acoustic model are detailed at 
\url{https://mfa-models.readthedocs.io/en/latest/acoustic/}.

All datasets have undergone substantial manual cleaning to improve data quality from the source releases, which we believe is crucial for making the pretrained models viable. The types of issues encountered include transcription errors, missing or incorrect speaker labels, extensive background noise, and various metadata, file format, and normalization errors. Each corpus required different preparation and cleaning, and the process was iterative and empirically driven: inspecting poor alignments, diagnosing errors, correcting issues across entire corpora, and repeating. This methodology motivated the development of dedicated corpus creation and correction tooling (Sec.~\ref{sec:support}). Still, given the scale of the data, some errors persist and many datasets contain other sources of noise like poor audio quality, which we mitigate through the structure of the training regime.

\subsubsection{Pronunciation dictionaries}
\label{sec:pron_dict}
MFA 1.0 contained dictionaries for 20 languages, which were sourced for most languages (besides English, French, German) from the GlobalPhone corpus \cite{schultz_globalphone_2014}, which provided only partial lexical coverage and contained transcription errors. For MFA 3.0, pronunciation dictionaries for the 20 core supported languages (Table \ref{tab:training_data}) have been rebuilt, now largely sourced from WikiPron \cite{lee_massively_2020}, which scrapes crowd-sourced pronunciations from Wiktionary and provides broader lexical coverage and narrower phonetic variation. Dictionaries were supplemented with G2P models trained on cleaned versions of the WikiPron dictionaries.  Rather than using a fixed language-specific phone set (such as ARPAbet for English as in MFA 1.0), MFA 3.0 uses a cross-linguistically harmonized narrow IPA phone set, cleaned, standardized, and extended with implementations of basic allophonic variation across languages\footnote{
\href{https://mfa-models.readthedocs.io/en/latest/mfa_phone_set.html}{https://mfa-models.readthedocs.io/en/latest/mfa\_phone\_set.html}. 
Mapping alignments to `broad transcription', e.g. the MFA phone set to ARPABET, is possible using `mfa remap alignments`.}, though an English ARPAbet model is still provided due to widespread usage. In total, MFA 3.0 provides dictionaries for 22 languages, with additional dialect dictionary support for 5 languages spanning 15 dialects (see Table 1). The  dictionaries for the remaining 17 languages are sourced from WikiPron and are limited in descriptions and availability of dialect forms. We treat the dialect as unspecified, but most forms are likely from standardized varieties.

Dictionary coverage is  extended to 34 additional languages through integration of the VoxCommunis pronunciation dictionaries \cite{ahn2022voxcommunis}, generated using the XPF  \cite{priva2021cross} and Epitran \cite{mortensen2018epitran} G2P systems, which MFA hosts to enable users to train aligners and align corpora in languages not covered by the core pretrained models. For out-of-vocabulary items across the 21 core languages, MFA includes G2P functionality based on a pair n-gram model trained from the pronunciation dictionary, using a weighted FST implementation of Phonetisaurus \cite{novak2016phonetisaurus} via  Pynini \cite{gorman2016pynini}.











\begin{table}[th]
  \caption{Training data differences for pretrained models in MFA 1.0 (left) and 3.0 (right). New dialects in 3.0 are underlined.}
  \label{tab:training_data}
  \centering
  \begin{tabular}{p{0.16\linewidth} r r r r p{0.2\linewidth}}
    \toprule
    \multicolumn{1}{c}{\textbf{Language}} & \multicolumn{2}{c}{\textbf{Hours}} & \multicolumn{2}{c}{\textbf{Speakers}} & \multicolumn{1}{c}{\textbf{Dialect}} \\
    \midrule
        Bulgarian & \multicolumn{1}{c@{\hspace*{\tabcolsep}\makebox[0pt]{$\rightarrow$}}}{8.0} &  35.1 & \multicolumn{1}{c@{\hspace*{\tabcolsep}\makebox[0pt]{$\rightarrow$}}}{77} & 172 & Unspecified \\
    Czech & \multicolumn{1}{c@{\hspace*{\tabcolsep}\makebox[0pt]{$\rightarrow$}}}{31.7} & 621.6 & \multicolumn{1}{c@{\hspace*{\tabcolsep}\makebox[0pt]{$\rightarrow$}}}{101} & 1.2k & Unspecified \\
    English & \multicolumn{1}{c@{\hspace*{\tabcolsep}\makebox[0pt]{$\rightarrow$}}}{982.1} & 3.5k & \multicolumn{1}{c@{\hspace*{\tabcolsep}\makebox[0pt]{$\rightarrow$}}}{2.4k} & 72k & {\scriptsize US, \underline{UK}, \underline{India}, \underline{Nigeria}, \underline{L2}} \\
    French & \multicolumn{1}{c@{\hspace*{\tabcolsep}\makebox[0pt]{$\rightarrow$}}}{27.0} & 1.0k & \multicolumn{1}{c@{\hspace*{\tabcolsep}\makebox[0pt]{$\rightarrow$}}}{100} & 18k & Unspecified \\
    German & \multicolumn{1}{c@{\hspace*{\tabcolsep}\makebox[0pt]{$\rightarrow$}}}{18.3} & 1.3k & \multicolumn{1}{c@{\hspace*{\tabcolsep}\makebox[0pt]{$\rightarrow$}}}{77} & 18k & Unspecified \\
    Hausa & \multicolumn{1}{c@{\hspace*{\tabcolsep}\makebox[0pt]{$\rightarrow$}}}{8.7} & 20.6 & \multicolumn{1}{c@{\hspace*{\tabcolsep}\makebox[0pt]{$\rightarrow$}}}{103} & 126 & Unspecified \\
    Japanese & \multicolumn{1}{c@{\hspace*{\tabcolsep}\makebox[0pt]{$\rightarrow$}}}{N/A} & 242.6 & \multicolumn{1}{c@{\hspace*{\tabcolsep}\makebox[0pt]{$\rightarrow$}}}{N/A} & 2k & Unspecified \\
    Korean & \multicolumn{1}{c@{\hspace*{\tabcolsep}\makebox[0pt]{$\rightarrow$}}}{20.9} & 99.7 & \multicolumn{1}{c@{\hspace*{\tabcolsep}\makebox[0pt]{$\rightarrow$}}}{99} & 301 & Unspecified \\
    Mandarin & \multicolumn{1}{c@{\hspace*{\tabcolsep}\makebox[0pt]{$\rightarrow$}}}{31.2} & 734.5 & \multicolumn{1}{c@{\hspace*{\tabcolsep}\makebox[0pt]{$\rightarrow$}}}{132} & 7.3k & {\scriptsize China, \underline{Taiwan}} \\
    Polish & \multicolumn{1}{c@{\hspace*{\tabcolsep}\makebox[0pt]{$\rightarrow$}}}{24.5} & 340.1 & \multicolumn{1}{c@{\hspace*{\tabcolsep}\makebox[0pt]{$\rightarrow$}}}{99} & 3.2k & Unspecified \\
    Portuguese & \multicolumn{1}{c@{\hspace*{\tabcolsep}\makebox[0pt]{$\rightarrow$}}}{26.3} & 783.1 & \multicolumn{1}{c@{\hspace*{\tabcolsep}\makebox[0pt]{$\rightarrow$}}}{102} & 2.9k & {\scriptsize Brazil, \underline{Portugal}} \\
    Russian & \multicolumn{1}{c@{\hspace*{\tabcolsep}\makebox[0pt]{$\rightarrow$}}}{26.5} & 443.1 & \multicolumn{1}{c@{\hspace*{\tabcolsep}\makebox[0pt]{$\rightarrow$}}}{115} & 3.2k & Unspecified \\
    Serbo-Croatian & \multicolumn{1}{c@{\hspace*{\tabcolsep}\makebox[0pt]{$\rightarrow$}}}{15.9} & 22.5 & \multicolumn{1}{c@{\hspace*{\tabcolsep}\makebox[0pt]{$\rightarrow$}}}{94} & 258 & Unspecified \\
    Spanish & \multicolumn{1}{c@{\hspace*{\tabcolsep}\makebox[0pt]{$\rightarrow$}}}{22.1} & 1.5k & \multicolumn{1}{c@{\hspace*{\tabcolsep}\makebox[0pt]{$\rightarrow$}}}{100} & 18k & {\scriptsize Latin-America, \underline{Spain}} \\
    Swahili & \multicolumn{1}{c@{\hspace*{\tabcolsep}\makebox[0pt]{$\rightarrow$}}}{11.1} & 434.5 & \multicolumn{1}{c@{\hspace*{\tabcolsep}\makebox[0pt]{$\rightarrow$}}}{70} & 1.2k & Unspecified \\
    Swedish & \multicolumn{1}{c@{\hspace*{\tabcolsep}\makebox[0pt]{$\rightarrow$}}}{21.7} & 439.0 & \multicolumn{1}{c@{\hspace*{\tabcolsep}\makebox[0pt]{$\rightarrow$}}}{98} & 1.6k & Unspecified \\
    Thai & \multicolumn{1}{c@{\hspace*{\tabcolsep}\makebox[0pt]{$\rightarrow$}}}{28.2} & 244.9 & \multicolumn{1}{c@{\hspace*{\tabcolsep}\makebox[0pt]{$\rightarrow$}}}{98} & 6.8k & Unspecified \\
    Turkish & \multicolumn{1}{c@{\hspace*{\tabcolsep}\makebox[0pt]{$\rightarrow$}}}{17.1} & 127.8 & \multicolumn{1}{c@{\hspace*{\tabcolsep}\makebox[0pt]{$\rightarrow$}}}{100} & 1.5k & Unspecified \\
    Ukrainian & \multicolumn{1}{c@{\hspace*{\tabcolsep}\makebox[0pt]{$\rightarrow$}}}{14.1} & 191.4 & \multicolumn{1}{c@{\hspace*{\tabcolsep}\makebox[0pt]{$\rightarrow$}}}{119} & 1.2k & Unspecified \\
    Vietnamese & \multicolumn{1}{c@{\hspace*{\tabcolsep}\makebox[0pt]{$\rightarrow$}}}{19.7} & 41.3 & \multicolumn{1}{c@{\hspace*{\tabcolsep}\makebox[0pt]{$\rightarrow$}}}{129} & 344 & {\scriptsize Hanoi, Ho Chi Minh City, Hue}  \\
    \bottomrule
  \end{tabular}

\end{table}

\subsubsection{Training regime}
\label{sec:training regime}

MFA 3.0 training builds from MFA 1.0 to more effectively use larger training datasets and integrates explicit pronunciation modeling. MFA uses a classical HMM-GMM architecture, adapted from Kaldi \cite{Povey_ASRU2011} recipes. MFA 1.0 used a three stage training pipeline \cite{mcauliffe_montreal_2017}: (1) monophone GMMs training; (2) triphone GMMs training to account for surrounding phonetic context; (3) speaker-adapted triphone training (SAT). In MFA 3.0, two additional training stages have been added. First an additional stage of triphone training using feature transforms from Linear Discriminant Analysis (LDA) has been added prior to the SAT stage. Second, MFA incorporates pronunciation and inter-word silence estimation into the lexicon \cite{chen2015pronunciation}.

The training regime has also been updated to leverage available training data while mitigating the impacts of less controlled data. The increased range of styles and speakers has been a net positive for modeling, but newer datasets often have quality issues. For example, CommonVoice is a crowd-sourced dataset and is especially liable to contain utterances with background noise or speech, which makes these recordings liable to introduce errors that propagate to later training stages. However, adding data into later training stages improves the performance of alignment on different styles of speech (e.g., conversational) and with variable environmental noise. 

As such, MFA 3.0 uses a data mixing strategy that incorporates noisier datasets progressively in the training cycles. For one initial training cycle (from monophone to SAT training) MFA uses clean and controlled corpora (like GlobalPhone \cite{schultz_globalphone_2013} or Multilingual LibriSpeech \cite{Pratap2020MLSAL}), and mixing in noisier corpora in iterations of SAT and pronunciation modeling (like the Multilingual TEDx corpora \cite{salesky2021mtedx}), and then using CommonVoice in the final round of training (see Table~\ref{tab:training_data}). For each of the training blocks, where relevant, each dialect is represented equally in the subset. Where equal distributions of dialects are not represented, then the underrepresented dialect is used maximally, and the other dialect(s) fill the rest of the subset.

\begin{table}[th]
  \caption{Training regime for MFA 3.0. Initial subsets only use read speech from clean corpora, with progressively noisier speech and recording conditions.}
  \label{tab:training_data}
  \centering
  \begin{tabular}{ l r r}
    \toprule
    \multicolumn{1}{c}{\textbf{Type}} &
                                         \multicolumn{1}{c}{\textbf{Utterances}} &
                                         \multicolumn{1}{c}{\textbf{Data}} \\
    \midrule
    Monophone & 10k & \multicolumn{1}{l}{Read}  \\
    Triphone & 20k & \multicolumn{1}{l}{Read}  \\
    LDA & 20k & \multicolumn{1}{l}{Read}  \\
    SAT & 20k & \multicolumn{1}{l}{Read}  \\
    SAT & 50k &  \multicolumn{1}{r}{+Spontaneous} \\
    Pronunciation probability & 50k &  \multicolumn{1}{r}{+Spontaneous} \\
    SAT & 150k &  \multicolumn{1}{r}{+Spontaneous} \\
    Pronunciation probability & 150k &  \multicolumn{1}{r}{+Spontaneous} \\
    SAT & All &  \multicolumn{1}{r}{+Noisy} \\
    \bottomrule
  \end{tabular}

\end{table}

For a subset of languages where noisy datasets such as CommonVoice constitute the primary training data (e.g. Russian, Czech, Spanish, Portuguese), we made manually corrected alignments to help guide training.  
Manually corrected alignments are used during training to constrain the alignments. The manual alignment is used to compare the predicted phone labels for a given frame to the ground truth label. When the ground truth label is not within the given set of predicted labels, the log-likelihoods associated with the predicted phones are overridden with a log-likelihood of -100,000.


MFA 3.0 incorporates explicit pronunciation probability modeling into the default training pipeline (``Pronunciation probability'' in Table~\ref{tab:training_data}), following the implementation of \cite{chen2015pronunciation} in Kaldi. After each SAT stage, pronunciation probabilities are estimated from the alignment lattices.
Silence probability and correction factors for preceding silence or non-silence are also estimated for each variant. These estimates are used to weight paths through the lexicon FST during subsequent training and alignment, making more frequent variants preferred. 

As an optional addition, MFA 3.0 supports the use of phonological rules to expand the pronunciation dictionary prior to training, generating additional variants wherever a specified rule can apply (e.g., flapping or cot-caught merger in American English). The goal is to more closely model the range of forms actually produced by speakers. In this case, the pronunciation probability stages of training additionally serve to prune the expanded dictionary: rule-generated variants that are unattested in the aligned training data are removed, while the original dictionary entries are retained regardless of attestation. 

Our experiments  evaluate the impact on performance of pronunciation probability modeling and phonological rules.

\subsection{Model adaptation and remapping}
\label{sec:adaptation}

One key feature in MFA 3.0 for improving alignments from pretrained models is acoustic model adaptation (\texttt{\footnotesize mfa adapt}).  
The pretrained model is first used to align the adaptation dataset, and the resulting alignments update the acoustic model's HMM statistics. Only the means of seen PDFs in the initial alignment pass are updated, not the variance, which limits the benefits of adaptation for varieties already well-represented in training data. For example, adapting MFA's English models to a data set containing General American English from adult male speakers results in fewer changes than adapting it to Scottish English from young children \cite{christodoulidou2025semi}.   MFA can also use manual alignments to guide adaptation  more precisely, which can outperform simply using more data, particularly for cases involving different speaking styles or speaker ages 
\cite{mcauliffegunter2025lsa}
.


MFA 3.0 has helper commands for remapping phone sets for dictionaries and alignment files.  The dictionary remapping command (\texttt{\footnotesize mfa remap dictionary}) allows for phone sets of one dictionary to be remapped to the phone set of a pretrained model for another language, as has been done for specific language pairs \cite{chodroff2025comparing}.  This functionality allows for the user to leverage large acoustic models like the Global English MFA model for languages that do not have enough data to train a dedicated model.  The alignment remapping command (\texttt{\footnotesize mfa remap alignments}) can convert the resulting alignments back to the original phone set for analysis.  The remapping may not be entirely lossless, however, as MFA allows many-to-one style mappings, such as representing a diphthong like /aj/ as a sequence of /a j/ that is supported in the pretrained model.

\subsection{Corpus creation and evaluation utilities}
\label{sec:support}




MFA 3.0 adds functionality to compare arbitrary alignments for the purposes of benchmarking. A common evaluation metric for forced aligners is recall and/or precision from the boundaries in the reference alignment and boundaries from a given aligner \cite{rehman2025bfa,zhu2022charsiu}.
A drawback of this method is that the two boundaries may not be separating the same phones, especially if the phone sets diverge significantly.


The evaluation algorithm for MFA uses a modified Levenshtein algorithm to align the two sets of phone intervals. Edit distance is a function of matching phone labels along with the interval's time points.  To mitigate the issue of different representations of phones across phone sets, users can specify a mapping between equivalent phone labels.  One-to-many mappings are also supported. Boundary error metrics are based on the aligned intervals.


MFA 3.0 has integrations with specialized libraries for expanded utility to support alignment, particularly with SpeechBrain \cite{ravanelli_speechbrain_2021}. Model hosting has migrated to HuggingFace Hub allowing for easier distribution and benchmarking of models with model cards \cite{mitchell2019model} detailing training data and configuration.

The first new utility is for segmenting longer audio files.  In addition to energy-based VAD algorithms, MFA 3.0 can leverage VAD models from SpeechBrain  to generate segments of speech.  With a pretrained model, dictionary, and transcript, MFA can iteratively segment the transcript as well.
Speaker diarization capabilities in MFA leverage speaker embeddings from SpeechBrain  to diarize speakers.
This is primarily useful for corpora with inconsistent speaker labels rather than as a general diarization solution (e.g.~\cite{bredin_pyannote_2020}).


Generating new transcripts from audio files can leverage WhisperX \cite{bain_whisperx_2023} or SpeechBrain.  These transcripts can be  used directly as input to alignment, or  checked against original transcripts to discover issues in the data.  MFA has the ability to use pretrained acoustic models, dictionaries and language models to generate transcripts as well, but modern ASR models like WhisperX and SpeechBrain have better performance.

Finally, there is support for language-specific tokenization, for 
languages that do not typically have spaces between words: Chinese (spacy-pkuseg: \cite{pkuseg}),  Japanese (Sudachi: \cite{takaoka2018sudachi}), Korean (Mecab-ko), and Thai (pythainlp \cite{pythainlp}). 



\subsection{Design and usability}
\label{sec:usability}

Alignment is typically one step in an analysis pipeline, rather than an end goal for users, and downstream considerations are often more important than performance benchmarks. MFA is designed to be easily installed and used either as standalone commands or as part of larger phonetic analysis pipelines (e.g. using LaBB-CAT \cite{fromont2012labb}, WebEMU \cite{winkelmann2017emu}, or PolyglotDB \cite{mcauliffe2017polyglot}). Three design principles have guided MFA development: sensible defaults with configurable parameters, extensive documentation, and human-in-the-loop iterative refinement.

An important principle of MFA development has been to have sensible defaults while exposing as many configurable parameters as possible. Aligner evaluations like the one in this paper (Sec.~\ref{sec:evaluation}) necessarily focus on default configurations, but configuration options can be difficult for end users to discover. We have iterated on the official documentation and tutorials with example data across multiple languages\footnote{
\href{https://montreal-forced-aligner.readthedocs.io/en/latest/first_steps}{https://montreal-forced-aligner.readthedocs.io/en/latest/first\_steps}
}, striving for ease-of-use and searchability, through workshops at various conferences and institutions 
\cite{mcauliffegunter2025lsa, mcauliffegunter2025mots}
. Other MFA users have put together a growing body of tutorials \cite{chodroff2021tutorial, xu2024gentle} that complements the academic literature on best practices for specific speech varieties and use cases (e.g.\ \cite{babinski2019robin, tosolini2025multilingual, christodoulidou2025semi,chodroff2025comparing}).

The training and alignment philosophy of MFA 3.0 emphasizes human-in-the-loop and iterative refinement of pretrained models and alignment outputs. While initial MFA outputs can be used as-is, MFA generates output files with  metrics for every aligned utterance that can be used for further inspection and correction. Aligners tend to be permissive and assume the transcript and pronunciations are correct. Violating this assumption in small ways (e.g., a word missing from the dictionary, wrong pronunciations, missing words in the transcript, etc) often only has local effects, as having all words specified in the dictionary is not a strict requirement for MFA. However, even small deviations can sometimes lead to catastrophic misalignments from the model. 
The per-utterance metrics help narrow down such issues and guide correction of transcripts or pronunciation dictionaries, and have been crucial in correcting and augmenting the training corpora and dictionaries used to develop MFA's pretrained models.

\section{Evaluation}
\label{sec:evaluation}

\subsection{Datasets}

Benchmark datasets were created from four corpora with manually corrected phone-level boundaries across three languages (Table~\ref{tab:datasets}).  The two English datasets are TIMIT \cite{garofolo_darpa_1993}, a corpus of read speech, and the Buckeye Corpus of spontaneous speech \cite{pitt_buckeye_2007}. The other two datasets represent Japanese and Korean, which are not typically included in benchmarks, but do have manually-corrected corpora of spontaneous speech, namely a subset of Corpus of Spontaneous Japanese (CSJ) \cite{maekawa_corpus_2003} and the Seoul Corpus \cite{yun_korean_2015}.  While TIMIT and Buckeye use a more narrow phone set, CSJ and the Seoul Corpus use a more broad phone set in their transcription.

For the spontaneous speech corpora, benchmark datasets were created by extracting utterances separated by 300 ms of silence, and padding utterances with 200 ms of silence on either side.  Utterances with three words or less or with unknown, cutoff, or excised words were filtered out.  For TIMIT, all utterances were included as is. 
For CSJ and the Seoul Corpus, no tokenization or splitting was done outside of the aligners.

\begin{table}[th]
  \caption{Summary of benchmark datasets}
  \label{tab:datasets}
  \centering
  \begin{tabular}{ l l r r r}
    \toprule
    \multicolumn{1}{c}{\textbf{Dataset}} &
                                         \multicolumn{1}{c}{\textbf{Language}} &
                                         \multicolumn{1}{c}{\textbf{Speakers}} &
                                         \multicolumn{1}{c}{\textbf{Hours}} &
                                         \multicolumn{1}{c}{\textbf{Phones}} \\
    \midrule
    TIMIT                       & English  & 630  & 5.38   &     216,284    \\
    Buckeye                       & English   &  40 & 17.12     &   703,336      \\
    CSJ      & Japanese  & 137  &  23.81  &   1,264,397 \\
    Seoul                    & Korean   & 40  & 26.99     &    1,161,552  \\
    \bottomrule
  \end{tabular}

\end{table}

\subsection{Alignment procedure}

We organize our evaluation around two goals. The first is to benchmark MFA 3.0 against other forced aligners out-of-the-box, to assess whether alignment performance has improved from MFA 1.0 and where MFA 3.0 stands relative to other current systems. The second is to evaluate the effect of specific functionality added in MFA 3.0: model adaptation, cross-language remapping, and the training options of pronunciation probability estimation and phonological rules.



For the first goal, the baseline alignments for each benchmark dataset are MFA 1.0 models alongside other forced alignment systems. The list of aligners includes other widely used systems (MAUS and SPPAS: \cite{bigi2012sppas,kisler2012signal}); language specific aligners for Japanese and Korean (Julius \cite{lee2001julius} and Korean Forced Aligner \cite{yoon2012forced});
 and recent neural alignment systems (Wav2Vec 2.0 \cite{baevski2020wav2vec}, WhisperX \cite{bain_whisperx_2023}, NeMo Forced Aligner \cite{rastorgueva_nemo_2023}, Charsiu \cite{zhu2022charsiu}, MAPS \cite{kelley2024mason}, and Bournemouth Forced Aligner \cite{rehman2025bfa}). MAUS and Korean Forced Aligner were accessed via their web interfaces; all other alignment runs were done via scripts.\footnote{
 \href{https://github.com/MontrealCorpusTools/mfa-interspeech2026}{https://github.com/MontrealCorpusTools/mfa-interspeech2026}
 } For TIMIT and Buckeye, MFA 3.0 was evaluated using both the Global English and US English ARPAbet pretrained models.\footnote{For the purposes of the benchmarking, the US English MFA dictionary was used to match the speech variety of TIMIT and Buckeye.} The Global English model was trained on 3.5k hours of speech with dialectal dictionaries, while the ARPAbet pretrained model was trained on 982 hours of LibriSpeech \cite{panayotov_librispeech_2015}, the same dataset as the original MFA 1.0 English model. For CSJ and the Seoul Corpus, the Japanese and Korean MFA pretrained models were used respectively. For all evaluation runs, G2P models were used to generate pronunciations for out of vocabularly (OOV) items.

For the second goal, we evaluate three aspects of MFA 3.0 functionality. To evaluate model adaptation, all pretrained models were adapted to each benchmark dataset (the +adapted rows in Tables~\ref{tab:phone_alignment_english}, \ref{tab:phone_alignment_csj}, and \ref{tab:phone_alignment_seoul}). To evaluate cross-language remapping for languages without a dedicated pretrained model, the Japanese and Korean MFA dictionaries were remapped to the Global English phone set, new G2P models were trained on the remapped dictionaries, and alignments were generated using both the default Global English pretrained model and an adapted version. Finally, to evaluate the effect of pronunciation probability estimation and phonological rules, new models were trained on each benchmark dataset under three configurations: the default training configuration, the default configuration with pronunciation probability estimation removed (-PP in Tables 5-7), and the default configuration with phonological rules added (+rules in in Tables 5-7).

\begin{table*}[!ht]
  \caption{Word alignment results for TIMIT and Buckeye.  * = neural aligner. Bold indicates best performance in column up to 0.25\%.}
  \label{tab:word_alignment}
  \centering
  \begin{tabular}{ l r r r r r r r r r r}
    \toprule
    & \multicolumn{5}{c}{\textbf{TIMIT}} & \multicolumn{5}{c}{\textbf{Buckeye}} \\
    \cmidrule(lr){2-6} \cmidrule(lr){7-11}
                       & Mean &
                       \multicolumn{1}{c}{\textbf{$t\leq10$}} &
                       \multicolumn{1}{c}{\textbf{$t\leq25$}} &
                       \multicolumn{1}{c}{\textbf{$t\leq50$}} &
                       \multicolumn{1}{c}{\textbf{$t\leq100$}} &
                       Mean &
                       \multicolumn{1}{c}{\textbf{$t\leq10$}} &
                       \multicolumn{1}{c}{\textbf{$t\leq25$}} &
                       \multicolumn{1}{c}{\textbf{$t\leq50$}} &
                       \multicolumn{1}{c}{\textbf{$t\leq100$}} \\
    \midrule
    \multicolumn{9}{l}{\textit{Baseline aligners}} \\
    MFA ARPA 1.0   & 28.18 & 35.07 & 56.95 & 84.03 & 95.76  & 27.22 & 39.30 & 61.99 & 87.91 & 95.65 \\
    MAUS           & \textbf{17.89} & 51.96 & 74.97 & 98.33 & {91.83}  & 32.78 & 42.50 & 62.15 & 83.77 & 93.44 \\
    SPPAS          & 31.23 & 29.68 & 52.49 & 82.82 & 95.37  & 38.63 & 33.60 & 56.54 & 81.79 & 91.36 \\
    Charsiu*        & 27.18 & 33.73 & 54.74 & 85.97 & 96.99  & 29.24 & 36.46 & 60.03 & 87.40 & 95.47 \\
    MAPS*           & 18.86 & \textbf{54.77} & \textbf{75.60} & \textbf{91.97} & {97.61}  & 54.44 & 38.24 & 55.07 & 71.96 & 79.06 \\
    BFA*            & 52.01 & 11.97 & 24.01 & 57.00 & 88.16  & 61.54 & 10.93 & 21.06 & 48.03 & 84.82 \\
    \midrule
    \multicolumn{9}{l}{\textit{ASR aligners}} \\
    MMS*            & 43.06 & 13.05 & 26.52 & 63.73 & 95.91  & 49.54 &  9.58 & 20.27 & 61.85 & 92.05 \\
    WhisperX*       & 110.04 &  1.95 &  4.21 & 15.55 & 53.98 & 110.90 &  1.31 &  2.85 & 13.48 & 57.38 \\
    NeMo*           & 78.24 &  7.61 & 15.50 & 38.23 & 70.03 & 88.62 &  7.00 & 13.31 & 35.81 & 63.09 \\
    \midrule
    \multicolumn{9}{l}{\textit{MFA 3.0 pretrained}} \\
    MFA ARPA 3.0              & 19.93 & 44.99 & 66.50 & 91.61 & \textbf{98.38} & \textbf{21.75} & \textbf{48.76} & \textbf{70.16} & \textbf{91.35} & \textbf{97.29} \\
    MFA Global 3.0            & 22.33 & 42.33 & 61.98 & 89.53 & 97.73 & 25.35 & 47.27 & 67.77 & 89.30 & 95.51 \\
    \bottomrule
  \end{tabular}
\end{table*}

Word alignment accuracy is calculated following \cite{rousso_tradition_2024} at thresholds of 10, 25, 50, and 100 ms;  for phone alignment accuracy 100 ms is omitted.   For both word and phone alignment we also report the mean boundary error.\footnote{For BFA, which inserts both onset and offset boundaries for each phone, we use only the onset boundary, as the inter-phone gaps in its alignments are generally an artifact of the CTC alignment algorithm rather than meaningful phone boundaries.}
In addition to the aligners evaluated for phone-level alignment, we include MMS \cite{pratap_scaling_2024}, WhisperX \cite{bain_whisperx_2023}, and NeMo \cite{rastorgueva_nemo_2023} as ASR-based neural aligners---systems that produce word-level alignment as a by-product of a neural ASR system with CTC loss---extending the comparison of \cite{rousso_tradition_2024}, who evaluated MFA against MMS and WhisperX.  Neural based word aligners generate beginning and end timestamps for all words, however, even for connected speech, the beginning of one word does not match the ending of the previous word, resulting in a choice of what to evaluate against.  Whereas \cite{rousso_tradition_2024} uses end timestamps, the current evaluation uses beginning timestamps for both word and phone boundaries for consistency, resulting in some differences in precise numbers.

Evaluating phone alignment accuracy is more difficult than word alignment, because while the sequence of words is the same across two aligners' outputs, the sequence of phones is not.  Evaluation practices vary widely \cite{mcauliffe_montreal_2017,kelley2024mason,rehman2025bfa}, making direct comparison across systems difficult. Aligner outputs were compared to reference alignments via \texttt{\footnotesize mfa compare\_alignments}, which calculates a Levenshtein alignment between the reference intervals from the benchmark and the aligned intervals from a given aligner. The cost function incorporates phone label accuracy and boundary distance. Per-aligner mapping files were used to establish equivalence in phone labels between aligners and benchmarks (i.e., ARPA ``T'' maps to TIMIT ``t''). 

For each boundary, the previous and following reference and hypothesis phones were collapsed into general manner categories (vowel, stop, approximant, nasal, fricative, sibilant/affricate), with silence and an ``unknown'' category for unknown words for a given aligner. To control for varying degrees of broad/narrow transcription across aligners and benchmarks, boundaries where the manner categories of the reference and hypothesis phones differed were excluded from evaluation. As an example, a boundary with a reference phone of ``aa'' would be included and compared to a boundary with a reference phone of ``EY1'', but not ``N''. This filtering affected the Buckeye corpus more than others (average of 908 boundaries removed across aligners, $<$1\% of the data). For TIMIT and Buckeye, flaps were treated as stops rather than approximants, and rhotics were collapsed with rhotic vowels (e.g. ``R'' and ``ER'' match to ``r'', ``axr'', ``er'' variants in TIMIT). This approach allows comparison across all aligners on a more level playing field, regardless of their phone set.

\section{Results}
\label{sec:results}

\subsection{Word alignment results}



Table~\ref{tab:word_alignment} shows word-level alignment accuracy on TIMIT and Buckeye.   MFA 3.0 substantially outperforms all three neural ASR-based aligners on both datasets, extending the findings of \cite{rousso_tradition_2024} to a larger comparison class. The gap is greatest for small thresholds (10, 25 ms) that are most relevant for speech research.  In comparison to other aligners, MFA 3.0 markedly outperforms MFA 1.0 on both datasets and ranks near the top compared to most other aligners. For Buckeye, MFA ARPA 3.0 and MFA Global 3.0 show the best performance, and all other aligners fall below even MFA 1.0 performance. On TIMIT, MFA performs third best, with MAUS and MAPS outperforming it. MAPS is a neural system designed for the forced alignment task rather than general ASR, suggesting better results for neural aligners when fine tuned for alignment as a task. However, other neural aligners (BFA and Charsiu) still do not reach the performance of classic systems. MFA performance is replicated in the phone-level results below.

\subsection{Phone alignment results}








\begin{table*}[ht]
  \caption{Phone alignment results for TIMIT and Buckeye. * = neural aligner. Bold indicates best performance in column up to 0.25\%.}
  \label{tab:phone_alignment_english}
  \centering
  \begin{tabular}{ l r r r r r r r r }
    \toprule
    & \multicolumn{4}{c}{\textbf{TIMIT}} & \multicolumn{4}{c}{\textbf{Buckeye}} \\
    \cmidrule(lr){2-5} \cmidrule(lr){6-9}
                       & Mean &
                       \multicolumn{1}{c}{\textbf{$t\leq10$}} &
                       \multicolumn{1}{c}{\textbf{$t\leq25$}} &
                       \multicolumn{1}{c}{\textbf{$t\leq50$}} &
                       Mean &
                       \multicolumn{1}{c}{\textbf{$t\leq10$}} &
                       \multicolumn{1}{c}{\textbf{$t\leq25$}} &
                       \multicolumn{1}{c}{\textbf{$t\leq50$}} \\
    \midrule
    \multicolumn{9}{l}{\textit{Baseline aligners}} \\
    MFA ARPA 1.0             & 16.38 & 48.96 & 76.22 & 95.02  & 17.58 & 51.08 & 76.24 & 94.48 \\
    MAUS                     & \textbf{11.26} & 63.55 & \textbf{86.76} & \textbf{97.82}  & 18.42 & 56.31 & 77.92 & 93.47 \\
    SPPAS                    & 21.44 & 35.29 & 64.31 & 92.25  & 26.14 & 40.97 & 67.60 & 89.59 \\
    Charsiu*                  & 17.79 & 40.96 & 69.73 & 95.46  & 18.43 & 42.83 & 71.67 & 95.22 \\
    MAPS*                     & \textbf{11.46} & \textbf{67.86} & \textbf{86.74} & 97.12  & 26.81 & 56.43 & 75.27 & 88.48 \\
    BFA*                      & 43.63 & 14.86 & 28.83 & 64.71  & 47.23 & 14.46 & 27.67 & 60.71 \\
    \midrule
    \multicolumn{9}{l}{\textit{MFA 3.0 pretrained}} \\
    MFA ARPA 3.0             & 12.11 & 61.85 & 83.56 & 97.40  & 13.87 & 62.93 & 83.53 & 96.04 \\
    \multicolumn{1}{r}{+adapted} & 12.21 & 61.70 & 83.42 & 97.26  & 13.78 & 62.80 & 83.54 & 96.09 \\
    MFA Global 3.0           & 12.36 & 64.08 & 83.11 & 96.80  & 14.97 & 62.61 & 81.51 & 95.23 \\
    \multicolumn{1}{r}{+adapted} & 12.47 & 63.94 & 83.05 & 96.66  & 14.93 & 62.14 & 81.46 & 95.27 \\
    \midrule
    \multicolumn{9}{l}{\textit{MFA 3.0 trained on dataset}} \\
    ARPA trained             & 11.95 & 63.61 & 85.26 & 97.12  & 13.82 & 60.23 & 82.80 & 96.30 \\
    \multicolumn{1}{r}{-PP}  & 14.02 & 61.89 & 83.23 & 95.11  & 13.93 & 60.24 & 82.61 & 96.21 \\
    \multicolumn{1}{r}{+rules} & 11.93 & 62.58 & 85.30 & 97.28  & 14.17 & 58.86 & 82.05 & 96.12 \\
    MFA trained              & 11.85 & 62.04 & 85.34 & 97.43  & 13.83 & 59.48 & 82.37 & \textbf{96.41} \\
    \multicolumn{1}{r}{-PP}  & 11.95 & 61.68 & 85.45 & 97.35  & 13.66 & 60.58 & 82.75 & 96.40 \\
    \multicolumn{1}{r}{+rules} & 13.09 & 61.64 & 83.99 & 96.34  & \textbf{12.90} & \textbf{63.77} & \textbf{84.36} & \textbf{96.66} \\
    \bottomrule
  \end{tabular}
\end{table*}

\subsubsection{MFA 3.0 versus baselines}

MFA 3.0 shows clear improvement over MFA 1.0 across all datasets. For English, mean boundary error drops from 16.4 ms to 12.1 ms on TIMIT and from 17.6 ms to 13.9 ms on Buckeye (ARPA phone set). For Korean, the improvement is larger still, from 20.7 ms to 14.8 ms. MFA 1.0 did not have a Japanese model, so no comparison is available.

Relative to other aligners, MFA 3.0 generally performs best across languages. As with the word-level results, MFA 3.0 (ARPA) outperforms all other aligners on Buckeye, including those that outperformed MFA 1.0. Again we see that MAUS and MAPS show the best performance on TIMIT with MFA following behind. SPPAS, Charsiu, and BFA all show much larger mean errors compared to any of the top three models. For CSJ and the Seoul corpus, the MFA 3.0 pretrained models outperform all other available aligners, including language-specific aligners (KFA and Julius).

\subsubsection{Model adaptation and phone remapping}

We evaluate two aspects of MFA 3.0 functionality for applying pretrained models to new data. The first is acoustic model adaptation, available for any language. The second is cross-language remapping combined with adaptation, evaluated for Japanese and Korean to simulate the case of aligning a language (or dialect) for which no dedicated MFA model exists.

For cross-language alignment, adaptation and remapping matter greatly. For both CSJ and the Seoul Corpus, aligning with the Global English MFA model using a remapped dictionary, without any adaptation, already performs competitively: for the Seoul Corpus, MFA English 3.0 outperforms KFA and BFA out of the box, and for CSJ it outperforms Julius and SPPAS, and falls just short of MAUS. Adapting the remapped model to the target language data significantly improves performance: for the Seoul Corpus, adapted MFA English 3.0 reduces mean error from 21.8 ms to 15.9 ms, closely approaching the performance of the dedicated Korean pretrained model; for CSJ, it improves from 14.3 ms to 11.7 ms, surpassing MAUS. 

For within-language adaptation, the effect is small across all datasets: adapting the pretrained ARPA or Global English model to TIMIT or Buckeye yields negligible changes in either direction, as does adapting the Japanese MFA model to CSJ. The large effect of adaptation seen in the cross-language case does not  carry over to within-language adaptation, even for spontaneous speech corpora like Buckeye and CSJ that differs from the largely read speech data in training.

\subsubsection{Pronunciation probabilities and phonological rules}

We evaluate the effect of two training options by comparing models trained on each benchmark dataset under the default configuration, with pronunciation probability estimation removed, and with phonological rules added (denoted ``-PP'' and ``+rules'' in the tables). 

Removing pronunciation probability estimation has little effect on most datasets: performance on Buckeye, CSJ, and the Seoul Corpus is essentially unchanged. The exception is TIMIT, where -PP results in notably worse performance for the ARPA phone set model (mean error increases from 12.0 ms to 14.0 ms), with a smaller effect for the MFA phone set model.
The effect of adding phonological rules (+rules) is more variable across datasets. On Buckeye, rules produce the largest improvement seen anywhere in our evaluation for this dataset, reducing mean error for the MFA phone set model from 13.8 ms to 12.9 ms and pushing it to the best result across all aligners on this dataset. On TIMIT and the Seoul Corpus, however, rules regress performance, and on CSJ the effect is negligible. The conditions under which rules help or hinder performance needs further experimentation that is outside of the scope of this paper, but we hypothesize these results may be a function of challenges in modeling pronunciation probabilities on smaller datasets.

\begin{table}[ht]
  \caption{Phone alignment results for CSJ.}
  \label{tab:phone_alignment_csj}
  \centering
  \begin{tabular}{ l r r r r }
    \toprule
                                          &
                                         Mean &
                                         \multicolumn{1}{c}{\textbf{$t\leq10$}} &
                                         \multicolumn{1}{c}{\textbf{$t\leq25$}}&
                                         \multicolumn{1}{c}{\textbf{$t\leq50$}} \\
    \midrule
        \multicolumn{5}{l}{\textit{Baseline aligners}} \\
                           MAUS  & 13.46 & 62.59 & 84.00 & 96.40          \\
                           Julius  & 19.25 & 39.02 & 70.64 & 95.35       \\
                           SPPAS & 17.75 & 44.43 & 75.17 & 95.53       \\ 
                           BFA*  & 78.44 & 2.19 & 4.90 & 24.69        \\ 
                   \midrule
                               \multicolumn{5}{l}{\textit{MFA 3.0 pretrained}} \\
                    MFA Japanese 3.0  & 10.82  & 63.44  & 88.34  & \textbf{98.81}   \\
                    \multicolumn{1}{r}{+adapted}  & 10.70  & 63.35  & \textbf{88.55}  & \textbf{98.90}   \\
                    MFA English 3.0   & 14.30  & 64.64  & 83.89  & 95.85   \\
                    \multicolumn{1}{r}{+adapted}   & 11.67  & 66.82  & 86.84  & 97.69   \\
                   \midrule
                              \multicolumn{5}{l}{\textit{MFA 3.0 trained on dataset}} \\
                    MFA trained  & \textbf{10.13}  & \textbf{67.75}  & \textbf{88.49}  & \textbf{98.80} \\
                    \multicolumn{1}{r}{-PP}   & \textbf{10.18}  & \textbf{67.56}  & \textbf{88.54}  & \textbf{98.81}   \\
                    \multicolumn{1}{r}{+rules}   & \textbf{10.26}  & 67.34  & \textbf{88.44}  & \textbf{98.77}   \\
    \bottomrule
  \end{tabular}

\end{table}


\begin{table}[ht]
  \caption{Phone alignment results for the Seoul Corpus.}
  \label{tab:phone_alignment_seoul}
  \centering
  \begin{tabular}{ l r r r r }
    \toprule
                                          &
                                         Mean &
                                         \multicolumn{1}{c}{\textbf{$t\leq10$}} &
                                         \multicolumn{1}{c}{\textbf{$t\leq25$}}&
                                         \multicolumn{1}{c}{\textbf{$t\leq50$}} \\
    \midrule
        \multicolumn{5}{l}{\textit{Baseline aligners}} \\
     MFA Korean 1.0  & 20.69 & 41.17 & 70.75 & 94.09 \\
                           KFA  & 22.34 & 52.51 & 74.80 & 91.47        \\
                           BFA*  & 85.81 & 1.96 & 4.76 & 23.17       \\ 
                   \midrule
                               \multicolumn{5}{l}{\textit{MFA 3.0 pretrained}} \\
                    MFA Korean 3.0  & {14.78}  & \textbf{61.11}  & \textbf{81.53}  & 95.70  \\
                    \multicolumn{1}{r}{+adapted}  & {14.60} & \textbf{61.17} & \textbf{81.77} & {95.89}   \\
                    MFA English 3.0   & 21.76 & 57.43 & 76.37 & 91.02   \\
                    \multicolumn{1}{r}{+adapted}   & 15.85 & \textbf{61.11} & 80.91 & 94.88   \\
                   \midrule
                              \multicolumn{5}{l}{\textit{MFA 3.0 trained on dataset}} \\
                    MFA trained  & \textbf{14.03}  & 58.05  & \textbf{81.73}  & \textbf{96.63} \\
                    \multicolumn{1}{r}{-PP}   & \textbf{14.24}  & 57.12  & 81.20  & \textbf{96.61}   \\
                    \multicolumn{1}{r}{+rules}   & {14.98}  & 56.08  & 79.12  & 95.96   \\
    \bottomrule
  \end{tabular}

\end{table}


\section{Discussion}

The goal of this paper was two-fold: 1) to evaluate MFA 3.0 performance against MFA 1.0 and other current aligners, and 2) to demonstrate available modeling utilities in MFA 3.0 that feed into end-to-end forced alignment for speech research. MFA 3.0 pretrained models demonstrate substantial improvements on benchmarking datasets compared to MFA 1.0 models, and state of the art performance across all three languages presented. Model adaptation and training configuration show mixed results against pretrained models, but provide key user-friendly flexibility for training and adapting models for their target data. In addition, MFA 3.0 provides a comprehensive evaluation toolkit that allows for felicitous comparisons of word/phone boundaries. Below we discuss the results in more detail, focusing on training utilities in MFA and the end user and challenges in benchmarking forced alignment.

MFA 3.0 outperforms MFA 1.0 and current aligners, with the exception of TIMIT, where MFA falls behind MAUS and MAPS. The effect of MFA 3.0's training regime on pretrained model performance can be seen in the differences between MFA 1.0 and 3.0 for the ARPA model, which is trained on the same data. The performance of MAPS is likely inflated due to training on TIMIT. MAUS performs the best on the read speech of TIMIT, but falls behind for spontaneous speech in Buckeye, suggesting that the training data for MAUS may be more tailored towards read speech. We also observe that neural alignment systems struggle to match the performance of HMM-GMM systems. However, one issue with evaluating WhisperX, NeMo, MMS, and BFA is that the alignment and timestamps they give for words/phones result in gaps between units without any interpolation or clear indication of which timestamp should be used for connected speech. There are differences in performance between selecting the end timestamp \cite{rousso_tradition_2024} versus beginning timestamp in this paper, suggesting that end timestamps might be better for certain systems. However, using end-time stamps alone does not improve their performance above MFA and other HMM-GMM systems.

MFA 3.0 train and adapt functionality do not show consistent gains over pretrained models. Training on the specific dataset results in the lowest alignment errors for MFA. However, the size and quality of the training dataset is crucial for good performance and may not be a viable option for most end users. Pretrained models and adapted remapped models approximate training well enough for most downstream analyses, removing the need for high quality/quantity data needed for training. Adaptation, however, will only yield substantial improvement where the target dataset is not well represented in the training data. The style and variety of speech in the benchmarks overlapped fully with the training data for MFA models (read and spontaneous speech from adult speakers of well-represented dialects). As such, MFA 3.0 pretrained models produce high fidelity alignments across these datasets without adaptation, and additional data of the same style minimally influences the model. Alternatively, we see adaptation improves alignment when the target data are used to adapt an out-of-language pretrained model (e.g., Japanese target and English model) with a remapped phone set. Adaptation has the largest effect when there is the most to adapt to \cite{christodoulidou2025semi}. While typical usage of MFA includes remapped phones \cite{chodroff2025comparing} and novel dialects \cite{coto2022computational}, they have not leveraged adaptation. Our results suggest that including model adaptation for these use cases is likely to improve alignment.

MFA development has explicitly focused on end users, particularly within language research domains. For an end user starting from clean audio and text in English, there are minimally two explicit components in the systems, mapping from words to a string of phones and mapping from audio frames to phones. However, there are several implicit assumptions that lay the foundation of these components that are often encoded into acoustic modeling with minimal visibility or control by users, despite forming the starting point of any downstream analysis. While this can provide users with easy-to-use tools, it constrains flexibility and obscures analytic decisions. For example, pronunciation modeling choices can influence not only boundary accuracy but also phone string accuracy in the resulting alignments. 
\begin{CJK}{UTF8}{min}
For aligners without pronunciation modeling, citation forms may not correspond to spontaneous speech forms (``for'' pronounced as ``F AO1 R'' or ``F ER0''), and even in English, homographs are common (i.e. ``read'' as ``R EH1 D'' or ``R IY1 D''), and even more impactful in a language like Japanese (i.e. ``一間'' romanized as ``hitoma'' or ``ikken''). 
\end{CJK}
This ignores valid variation in speech that not only impacts boundary alignment, but introduces errors that can impact downstream analysis for researchers (e.g., vowel formants).
Often end users are stuck with the choices made by the developer(s). MFA prioritizes ease of installation and use, but provides a toolbox of commands for experimenting with different models, dictionaries, training, and adaptation configurations along with evaluation tools, so the end user does not have to rely on invalid assumptions for their data.

Due to the complexities of modeling choices and dataset construction, felicitous comparisons of alignments are difficult and researchers take varied approaches to evaluation. The primary distinction between methodologies is whether to consider the predicted phone label when evaluating boundary accuracy. Some evaluation methods ignore the phone label accuracy and take a naive approach to calculate precision and recall metrics based on the closest boundary\cite{rehman2025bfa}, or based on whether the midpoint of the reference interval is contained by the hypothesis interval\cite{mahr2021performance}. Assuming the phone sequence and number of boundaries are the same in hypothesis and reference alignments, this method is valid. In practice, however, this approach is invalidated by implicit assumptions in the models. For example, if there is a gap between phone boundaries (e.g., BFA), there is effectively double the number of possible boundaries for comparison which inflates accuracy metrics. Other approaches attempt to include phone label accuracy alongside boundary precision. However, evaluation with phone labels is not straightforward due to differences in the phone sets of reference alignments and acoustic models. Narrowly transcribed corpora, like TIMIT and Buckeye, will be mapped against broader phone sets in the aligners. As such, some evaluations restrict the set of evaluated words \cite{mcauliffe_montreal_2017} or skip the pronunciation look up entirely and train/evaluate only on oracle phone strings \cite{kelley2024mason}. The approach in MFA's alignment evaluation utility provides a balance between naive approaches and stricter methods while allowing for felicitous comparisons of boundaries.

Overall, our goal is to make a wide range of utilities available to support speech researchers use cases. Benchmarks provide a good picture of the overall performance of MFA. Equally importantly, the functionality provided in MFA 3.0 provides users with the ability to tailor MFA to their specific needs.

\section{Conclusion}

We have presented key updates to the Montreal Forced Aligner that have improved performance on benchmarks across three languages, along with summarizing new utilities included in MFA 3.0. MFA 3.0 demonstrates state of the art performance against existing aligners and provides users additional functionality to tailor MFA to their own data. Future development of MFA will continue to expand utilities to support alignment outside of well-represented speech varieties and populations.


\section{Acknowledgments}

We acknowledge funding from SSHRC \#430-2014-00018, FRQSC \#183356, CFI \#32451 and SSHRC CRC program to Morgan Sonderegger; SSHRC \#435-2014-1504 and the SSHRC CRC program to Michael Wagner; and NIH \#R01DC019645-03 awarded to Katherine C. Hustad.

\section{Generative AI Use Disclosure}

No generative AI was used in preparing this manuscript.

\bibliographystyle{IEEEtran}
\bibliography{mybib}

\end{document}